\def\year{2023}\relax
\documentclass[letterpaper]{article} 
\usepackage{aaai23}  
\usepackage{times}  
\usepackage{helvet}  
\usepackage{courier}  
\usepackage[hyphens]{url}  
\usepackage{graphicx} 
\usepackage{natbib}  
\usepackage{caption} 
\DeclareCaptionStyle{ruled}{labelfont=normalfont,labelsep=colon,strut=off} 
\usepackage{algorithm}
\usepackage{algorithmic}

\usepackage{newfloat}
\usepackage{listings}
\floatstyle{ruled}
\newfloat{listing}{tb}{lst}{}
\floatname{listing}{Listing}
\usepackage{subcaption}
\usepackage{amsthm}

\theoremstyle{definition}

\title{Neural-Base Music Generation for Intelligence Duplication}
\author{
    Jacob E. Galajda\\
    Kien A. Hua
}
\affiliations{
    University of Central Florida\\

    4000 Central Florida Blvd \\
    Orlando, FL 32816, United States \\

    Jacob.Galajda@ucf.edu
}

\begin{document}

\maketitle

\begin{abstract}

There are two aspects of machine learning and artificial intelligence: (1) interpreting information, and (2) inventing new useful information. Much advance has been made for (1) with a focus on pattern recognition techniques (e.g., interpreting visual data). This paper focuses on (2) with intelligent duplication (ID) for invention. We explore the possibility of learning a specific individual's creative reasoning in order to leverage the learned expertise and talent to invent new information. More specifically, we employ a deep learning system to learn from the great composer Beethoven and capture his composition ability in a hash-based knowledge base. This new form of knowledge base provides a reasoning facility to drive the music composition through a novel music generation method.

\end{abstract}

\section{Introduction}

Machine and deep learning applications are common solutions to multimedia automation. This often involves mimicking human behaviors to achieve systems that can, for example, detect if someone is running or jumping if given an image or piece of video. These patterns can be easily detected by humans and thus we have learned how to automate these criterion with a series learning techniques on a dataset composed of many samples. However, we cannot readily use these techniques to learn human thought such as a particular style of a single individual's art, work, or craft; as mental patterns are abstract, not visible, and less detectable by humans and systems alike. 

The invention of many pattern recognition techniques capable of learning universal human ability has enabled numerous important automation applications in recent decades.  The new ability of learning the expertise of a specific talented individual can take automation systems to the next level to support much more sophisticate applications.  Such a system needs to learn the thinking process of the expert but also apply the learned knowledge to create new solutions for new situations. The challenge is twofold: (i) How to capture and store the expertise, and (ii) what mechanism can reference the stored knowledge to create new solutions?  We call this direction of AI research \emph{Intelligence Duplication} (ID).  In this initial attempt, we focus on studying ID through music composition.  Our goal is to revive a great composer such as Beethoven to create his new music for us today.



To elaborate ID further, we can equate the traditional artificial intelligence workflow to the temporal and occipital lobes of the human brain. The temporal lobe is used to understand hearing, memory and language based tasks whereas the occipital lobe is responsible for interpreting visual information. Regardless of the specific neurons being activated in each of our brains, we each can read the same piece of literature and comprehend what the text contains with little to no difference across each other. On the contrast, we would each interpret the literature's deeper meaning using our own creativity and reasoning within the frontal lobe of our brains. In this situation, it would be statistically unlikely for two individuals to share the exact same interpretation of a literary work -let alone sharing the exact same neural architecture.

Deriving an equivalent neural architecture of one's frontal lobe could theoretically give us the ability to use one's creative reasoning in a variety of automated applications. Systems attempting to have stylized generation should consider this in their architectures -otherwise the style can easily be lost as the system learns its own unique style asides from the dataset's composer. In this paper, we discuss how current solutions to automated learning are not sufficient for these tasks. As modern deep learning solutions do not focus on frontal lobe tasks, we must seek a new methodology for \textit{intelligence duplication}. Given these constraints, our contributions are as follows:

\begin{itemize}
    \item We propose and implement a knowledge-base technique for intelligence duplication
    \item We demonstrate a methodology for authentication of the synthetic solutions. 
\end{itemize}

These contributions will provide a clear benchmark for future ID research applications.

The remainder of this paper is organized as follows.  We discuss related works in Section 2.  The challenges of ID research is examined in Section 3.  In Section 4, we introduce Neural Base, our knowledge-base approach to ID.  The results of our performance study are presented in Section 5.  Finally, we conclude this current work and discuss our future research on ID.

\section{Related Works}

Our proposed solution to intelligence duplication captures the knowledge representations in a novel knowledge base application called Neural Base; and the invention process is based on a search technique designed for the Neural Base for music composition.  We discuss some related works in this section.

\subsection{Knowledge Base Applications}

Knowledge bases are a collection of rules generated through user experience or algorithmic expertise. They allow for systems to be automated on certain criterion being met, much like a long series of conditional logic in structural or object-oriented programming. The knowledge base can be used to derive context in a variety of domains, including literature and various multimedia \cite{brezillon1999context, ishikawa1993model}. This technology was popularized in the advent of AI, however, was being phased out with the increasingly powerful computing machines that have enabled deep learning technology.

While much of AI research today has been directed towards machine and deep learning, knowledge base systems are still used to this day. In recent cases, knowledge bases are used to power applications like chat-bots \cite{ait2020kbot} and client knowledge-base systems \cite{yau2021artificial}. These systems are powerful because they act like a relational database and integrate very closely with information that is commonly stored in this format.

This scheme could potentially allow for us to capture all of the rules necessary for a certain application as long as we are able to obtain each and every rule associated with the given tasks.  However, it is quite evident that a conventional knowledge base system will require a massive amount of refinement and rule sets in order to accomplish this. If we were to even consider this approach for intelligent duplication, we must question how we could automatically generate the rules an individual expert would have for a specific task.  

\subsection{Music Generation}

We chose to facilitate this theory through music generation research as music is one of the most popular common languages across cultures and offers complex mental patterns that are quite difficult to recognize and replicate. Much of the music generation research field involves one of two methods for creating new pieces: note generation and segment concatenation.

\subsubsection{Note-Generative Models}

On the one side, note-generative models like the SeqGAN \cite{SeqGAN2017} offer much versatility in output. These systems use measures of a song for training, however, the generation will only take place at the note-level -utilizing a modified Generative-Adversarial Neural Network (GAN). This gives the system more opportunity for new compositions and greater freedom of artistic ability \cite{mogren2016c,kulkarni2019survey}. This approach is not without it's flaws; as the increased freedom in generating new songs often leads to \textit{experimental} compositions that do not contain a well-definable structure.

In other multimedia experiments using the Generative-Adversarial Neural Networks, \cite{ganDisadvantages} noted that the efficiency of the GAN relied on the use of synthetic training data and the ability to reduce the complexity of the images. This allows for the generated images to have similar quality while retaining minimal loss. However, these smaller details are often what define the uniqueness of a song and could be neglected.

It becomes extraordinarily difficult to learn compositional style on a note-by-note case due to the input size. As there are many ways to create a measure of a song, composers do not intrinsically select the next note to place by only analyzing the previous note. Thus, learning how composers place measures (or segments) together is more intuitive for our purposes.

\subsubsection{Segment-Concatenative Models}

On the other side, segment concatenative (or measure-concatenative) models utilize the relationship between two adjacent measures and can create new compositions \cite{BiLSTMMusic2019}. By utilizing a neural network schema such as the Long-Short Term Memory (LSTM) on both ends of the network, two segments can be fed into the system at the same time. The cross-entropy loss between these networks will be used in backpropagation, thus training the system on the loss between two segments.

As music generation research has evolved, stylized generation has become a more popular topic. One attempt at stylized generation, MuseNet \cite{Toprceanu2014MuSeNetCI}, utilizes a sparse transformer neural network to retain the MIDI input received in training. The neural network itself behaves much like a sophisticated Long-Short Term Memory (LSTM) neural network, in that it can take a variable size input denoted by time steps and process a corresponding encoding for each measure or \textit{token} of a song. This allows the model to maintain song structure quite well.

The researchers describe this model as using MIDI tokens, in other words a recording of an event in a song. Each token contains the name of the composer as well as a corresponding audio clip from the song. This allows the system to generate in a particular style, however, a prompt must be loaded into the system to initialize. Additionally, much emphasis in maintaining the style throughout generation is within the large transformer neural network resulting in strain placed on the neural network that can inhibit the system's ability to learn and retain style from an individual composer when introduced to many training samples. 

\subsubsection{Copyright Law Surrounding Generated Music}

Some issues presented to segmented concatenative models pertains to the copyright laws surrounding modern music. While using long segments has shown to help with reducing the system loss, as in DeepHear \cite{LongerMeasuresBriot2020}, the U.S. Copyright Office states that a song can be composed into two domains, sound recordings and the musical work itself \cite{copyrightWhatMusicians}. Having the segments be small enough to be considered sound recordings offers a legal benefit to music generation research, especially as cases like Queen and David Bowie v. Vanilla Ice \cite{turville2017emulating} were fought over simple changes in the drumbeat.

If we compare music copyright to literature and company taglines, no entity can own the exclusive rights to one or two common words. It increasingly becomes more difficult to find copyright issues by sampling from a variety of songs and sources. Of course, this issue is entirely mitigated by using songs dedicated to the public domain, as they do not fall under copyright law. This is an important consideration for modern music generation research we must recognize.

\section{Intelligent Duplication (ID)}

To our knowledge, current state-of-the-art music generation systems are not equipped to learn the unique compositional style of an artist and create new songs in the given style. As stated by the researchers \cite{trost2014getting,li2022research}, their goal is to optimize \textit{song pleasantness} and creating new songs that sound human-like. Additionally, these models train on large datasets of similar music in order for the model to learn what sounds pleasant. In the following sections, we will formally describe the differences between traditional AI and our Neural Base approach to Intelligence Duplication.


\subsection{Separation From Traditional AI}

Our objective in this paper is to design, implement, and test a new paradigm of artificial intelligence known as \textbf{Intelligence Duplication} (ID). For our purposes, we facilitate this in the application of music generation. In contrast to music generation research, our primary objective in the required neural network training is to extract the reasoning from the artist's frontal lobe through their works and store the learned information is a knowledge base called the Neural Base. Thus, our generated Neural Base is optimized for intelligence duplication; not just song pleasantness like other music generation platforms. Much like traditional knowledge bases, the Neural Base is a collection of the learned rules needed to make decisions that replicate the dataset composer in creating new music.

\begin{figure}
     \centering
     \begin{subfigure}[b]{0.45\textwidth}
         \centering
         \includegraphics[width=.88\textwidth,keepaspectratio]{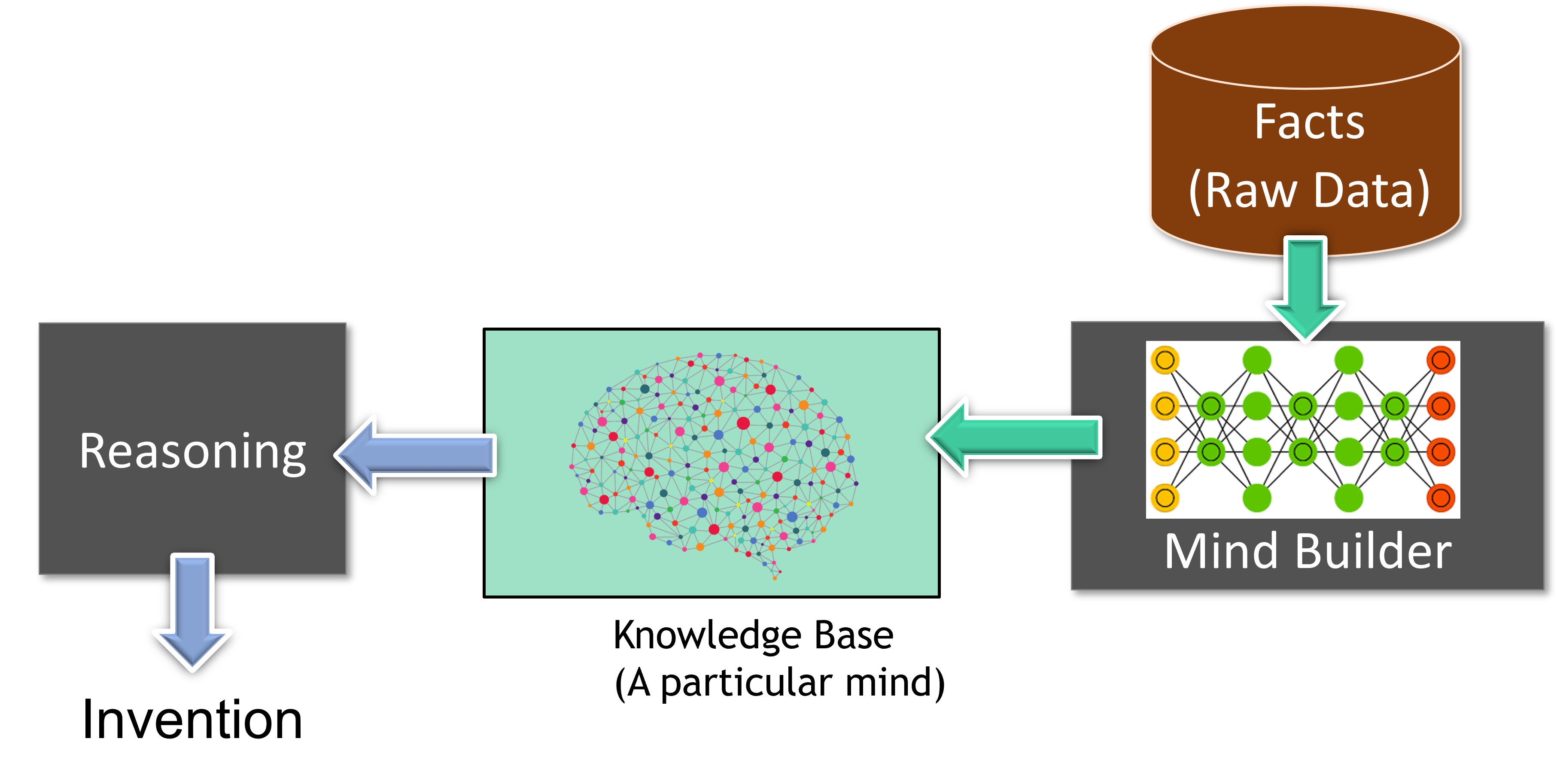}
         \caption{}
         \label{fig:Step1}
     \end{subfigure}
     \hfill
     \begin{subfigure}[b]{0.45\textwidth}
         \centering
         \includegraphics[width=.88\textwidth,keepaspectratio]{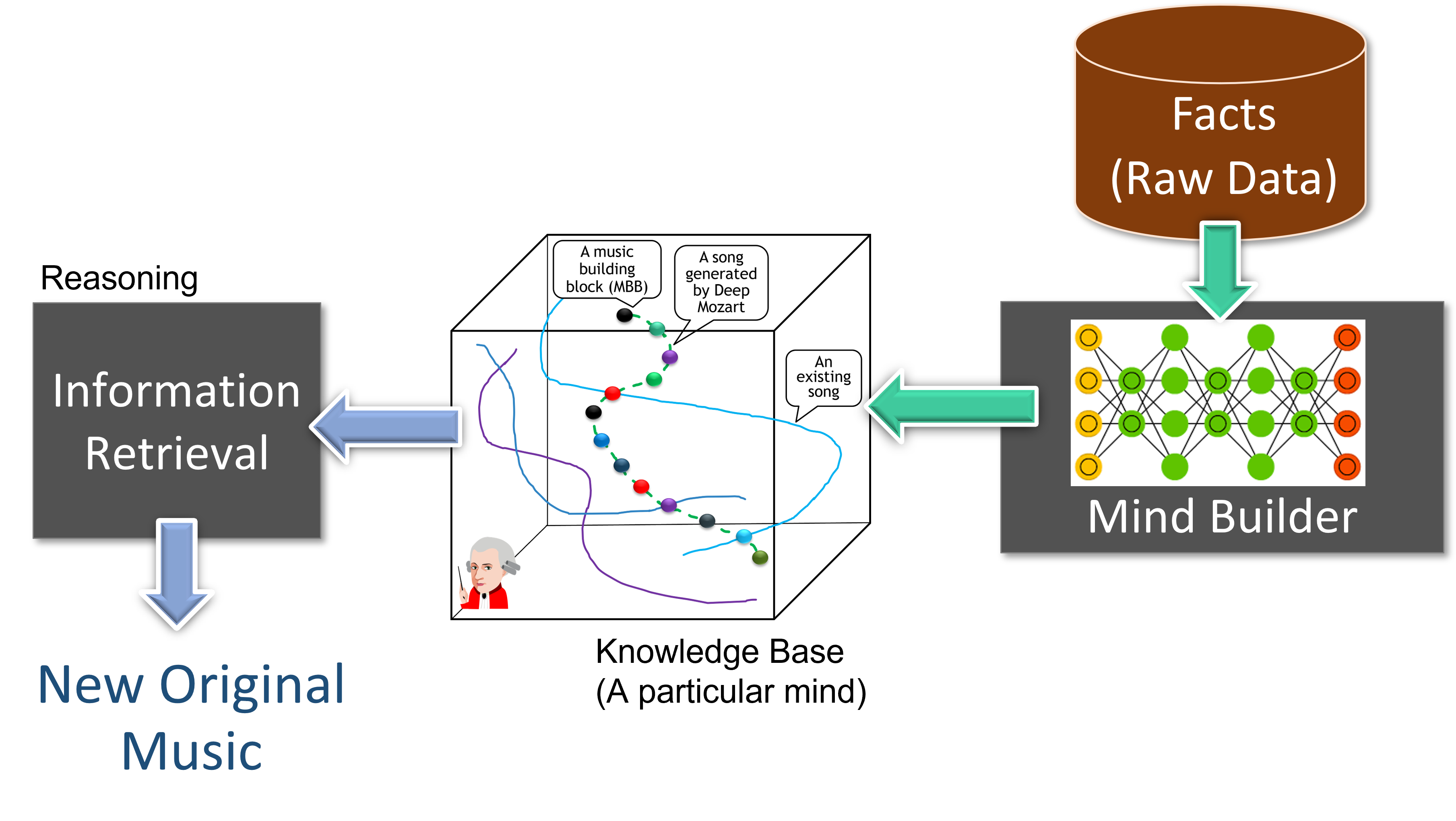}
         \caption{}
         \label{fig:Step2}
     \end{subfigure}
     \caption{Intelligence Duplication research targets localized expertise learning (Figure \ref{fig:Step1}) and utilizes this mental process in Figure \ref{fig:Step2}.}
     \label{fig:IDResearch}
\end{figure}

Analyzing  Figure \ref{fig:IDResearch}, we can obtain a general blueprint for ID applications. Applying the Neural Base in the form of a frozen neural network, we can use the inference from the \textit{mind} to conduct a query based off of learned composability between segments. As with deep learning solutions, the neural network would generate new segments of data to be concatenated together. In our approach, we employ the Neural Base to generate \textit{the address} of the next segment from within the Neural Base in the form of dynamic hash encodings generated from a neural network component. This is a new approach of inference using a knowledge base, different from the conventional conditional logic inference mechanism. 

With this considered, we believe ID can be achieved if we design our neural network to optimize the learning of segment placement as specified by the original composer. This will allow us to create new songs that retain the same compositional style with our novel Neural Base approach.

\subsection{Challenges}

This new research paradigm proposes a unique requirement on training datasets. Rather than incorporating more samples into the dataset for better performing neural networks, we must \textit{reduce the dataset} to include samples exclusively from our subject. This stringent requirement is quite counter-intuitive by comparison to traditional AI techniques. However, it is a necessary requirement for ID research.

ID research focuses neural network training to learn habits of an individual. By adding more origins (or composers in our case) to the dataset, multiple sources of reason will affect the neural network's ability to extract the reasoning of our intended target. Thus, it is imperative to use as many samples from the targeted composer.

For our case, we are limited to the samples generated from one composer exclusively. This can become an issue when we need to learn complex patterns and generate new songs, a phenomena noted as segment sparsity \cite{segmentSparsity2021}. This occurs when there are not enough composable segments when generating new songs. To remediate this, \cite{ucfRevivingMozart} propose synthetically generating new segments, however, they noted that new segments must be placed in accordance with an existing song. Otherwise, generated segments will disrupt the original dataset. For example, if random measures of Mozart were added into an existing work of his, they can sound out of place if no structure is provided. This issue becomes more prevalent as more segments are added, thus decreasing the performance.

Our solution to this problem does not involve adding more synthetically generated segments into the training dataset. As the timing issues involved with this approach can disrupt the neural network training, we propose an unorthodox technique to remediate segment sparsity. In this paper, we outline our neural-base application that can overcome limited training data presented with intelligence duplication research.

\section{Neural Base Implementation}

In order to construct our Neural Base application, we must reference Figure \ref{fig:IDResearch} first. The most complex part of this research is the knowledge extraction. To do this, we implement a bidirectional neural network \cite{schuster1997bidirectional} that reads in two adjacent segments concurrently.

We implement our information retrieval task as a simple heuristic application that makes retrievals within the Neural Base using the previous input. When generating a new song, we randomly select a position within the dataset and query against the Neural Base to select the next song.

\subsection{Neural Network Architecture}

In order to facilitate ID research, we must ensure that our neural network learns the positioning style of our composer. In traditional Computer Vision tasks, feature extraction plays an important role for eliminating redundant or useless features in hopes to optimize network performance. In the case of ID, we do not know exactly what features, say Mozart or Beethoven, would have used to compose a work. Thus, incorporating all available features into the input layer will give us the closet possibility of accuracy.

To reduce the complexity, we opt for a bidirectional Long-Short Term Memory (LSTM) neural network architecture with hidden layers to aggregate a softmax hash encoding for each LSTM. The strength in the LSTMs lie in the ability to read in variable length input. This allows for us to rapidly experiment with segment length as well as trim final segments in a song without padding the final song.

With the introduction and adaptation of the Transformer neural network \cite{originalTransformerPaper} in many aspects of multimedia AI, the LSTMs can be substituted with a corresponding forward and backwards Transformer neural network as they have similar properties. In future works, we intend to compare this model with the Bidirectional LSTM to see if there is a performance difference.

\begin{figure}
    \centering
    \includegraphics[width=.6\columnwidth,keepaspectratio]{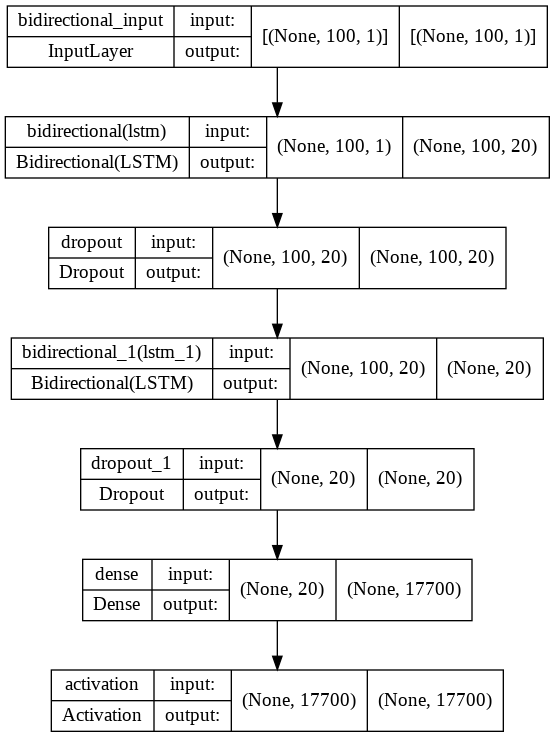}
    \caption{Two layers are defined for the forward and backwards direction. Each LSTM passes through additional layers. The output from both LSTMs are then evaluated using a cross-entropy loss.}
    \label{fig:network}
\end{figure}

\section{Performance Evaluation}

Many standard performance evaluation techniques (e.g. precision, recall, etc.) are available for studying the effectiveness of pattern recognition methods. Evaluation of intelligence duplication requires new performance evaluation methodology that authenticates the generated pieces. In this section, we introduce novel assessment techniques and discuss our performance results based on this proposed methodology. We propose two methodologies to test our system, one mapping heuristic information from the dataset and system output and the other incorporating a user-feedback survey to determine the authenticity of each piece.

The purpose for the mapping our output alongside the dataset is to heuristically compute how \textit{close} our output is to relative samples in the dataset using sufficient techniques. The visualization will help us to make relative assessments of our system's performance and establish a benchmark for future comparisons with revisions or even other systems.

The intent behind the survey is to gauge how well our music performed in comparison to the masterpieces we utilize in the dataset. We ask users to determine if the sample is human or computer composed. This reduces the overall survey complexity, in that our objective is on intelligence duplication and not song pleasantness. However, samples from our neural network and survey are included in the supplementary materials. Generally speaking, pleasantness is a metric commonly used to assess the subjective quality of a song. We expand further on this definition in the survey section.

\subsection{Network Performance}

Before we begin discussing the performance evaluation, we must formally present the dataset, network architecture and training parameters. It should be noted that the computationally expensive operations, like training and generation, were carried out on an Ubuntu server. This server has 128GB of RAM, an AMD Ryzen\texttrademark Threadripper\texttrademark CPU and 2 Nvidia GTX Titan GPUs. Further details to the system performance will be described in each corresponding section. Tasks that were not computationally expensive, like graphing and data preprocessing were carried out on a Dell G7 with 16 GB of RAM, an Intel 8$^{th}$ Generation i7 CPU and Nvidia GeForce GTX 1080. Similar or better computer architectures should be able to reproduce our experiments, however, there may be increased runtimes with smaller systems.

\subsubsection{Dataset and Preprocessing}

The dataset used in our experiments was derived from the popular AI challenge site, Kaggle\footnote{Dataset: https://www.kaggle.com/datasets/soumikrakshit/ classical-music-midi}. Additional Beethoven samples were collected from Classical Archives. This allowed for us to obtain enough information for our training as well as provide the Neural Base with enough samples from other composers to generate more pleasant sounding music.

The dataset contains MIDI information (denoted in .mid or .midi format; .mid preceding .midi format). Each file contains \textit{MIDI events} which include notes, chords, rests, etc. For the purposes of the neural network training, we utilize the popular Python module, Music21 \cite{mitMusic21Toolkit} for interpreting these events and to provide us with proper segmentation by each measure.

For the task of neural network training and song generation, we extract and present the MIDI information with integer encodings. This format is more intuitive for the neural network to train with. Segments are arbitrarily divided into 100 timesteps per measure, however, this value can be modified to have smaller or larger segments. This value is roughly less than one measure long when considering the tempo and time signature of each sample.

We extracted the note and chord information from each segment and one-hot encoded this information. This offers the most benefit in categorical crossentropy loss as well as reducing the network strain with in the final layer to compute the results. The total dataset composition consists of 292 classical piano sontas from 15 different classical composers, including Beethoven. 29 of these songs were from Beethoven while the rest were distributed across the remaining composers. To train the neural network, the 29 Beethoven songs were considered, however, the full dataset was used for generating new songs.

In the case of our graphing task, an entirely different representation for our graphing dataset is considered -even though we utilize the same MIDI information from our training dataset. Utilizing the Comma Separated Values (.csv) file format, we can represent our information much like a relational database for the graphing software. This, in turn, allows for us to streamline our dataset comparisons. In order to denote the origin of each song, we can specify the color of each entry accordingly. The only distinction to be made between each dataset is the graphing dataset will also contain the output from our completed system for comparison. The code to achieve this will be included in the supplemental materials.

\subsubsection{Neural Network Training}

The neural network architecture described in Figure \ref{fig:network} is generated directly from the Keras platform \cite{chollet2015keras}. This platform extends Tensorflow and provides an intuitive interface for researchers to quickly develop deep learning solutions. The inputs to both Bidirectional LSTM take in an adjacent pair of measures, as created in the data preprocessing segment. Once fed through, we optimize the output and take the cross entropy loss between both outputs and work to reduce this loss. The system was set to run for 200 epochs, and achieved a loss of 0.0243 for our experiment -which is impressive considering the lack of training data. Since we will be comparing our iteration to one without the extended dataset, we only needed one iteration of this neural network, thus this factor would remain constant throughout both iterations.


On multiple attempts to optimize the individual layers, the model consistently ran under 2 hours on the previously described computer architecture. Reducing the number of layers and neurons required tended to increase the speed of training, however, the ability to learn was slightly impacted due to the smaller dataset. We also conclude that modifications to this architecture like adding additional LSTM layers did not always result in better learning. In many cases, this would dramatically increase the time to complete training but would often result in decreased performance.

\subsubsection{Generation Through Retrieval}

Once the training is complete on the single-source dataset, we will have extracted and formed the Neural Base. Given that we are only interested in the individual building blocks of a song and how they are placed together, this Neural Base is the representation of how a composer would position various measures together.

We theorize that adding new segments with valid positioning into this Neural Base would provide us the means of generating new and refined samples in the same compositional style of the initial Neural Base. In turn, this would make songs more pleasant as there are more options for our Neural Base to select from. To achieve this, we can add samples from other composers into the dataset. This will have no impact on the established Neural Base rules while allowing the Neural Base to make decisions on similar data because the neural network component will not learn from new data.

We can randomly select a segment from within the dataset and query this into the hash network, which facilitates the new form of inference for our Neural Base. The result of the query provides the next candidate segments for us to select into our generated song. This process can repeat by taking the previous selection and using it as input into the next query. Through a series of hash network retrievals, we can obtain a new song that retains the same compositional style learned in training.

\subsection{Graphing Our Results}

In order to process music information into a graphical format, we must first prioritize MIDI events that will accurately depict key components of the song. According to previous research \cite{walker2003sonification,kroonenberg2021musical}, elements such as timbre (note quality), pitch change/melodic interval (time in between two notes), pan and volume play a key role in graphing music. Since the MIDI file format is a digital music sheet, volume remains constant. This would not be the case if we were processing .mp3 or .wav files. Additionally, the MIDI samples we have do not pan like their waveform counterparts. For these reasons, we only considered the changes within the note intensity and pitch with respect to time. This gave us the ability to graph a 3D line of each song depicting key features with the progression of time.

Music21 \cite{mitMusic21Toolkit} provides us with many tools we can use to efficiently prune relevant features and compose a reduced feature subset. We can extract features that encompass both note density and pitch changes throughout each measure and correlate this change using matplotlib -a popular Python graphing module \cite{Hunter:2007}. The source code to our graphing program is included as supplementary materials to the submission.

We extracted 30 features from each song and stored the result in the form of a Comma Separated Value file. Using the Pandas Data Science Library \cite{mckinney-proc-scipy-2010}, we were able to create two separate categories to compose our x and y axes, respectfully. The x axis will primarily correspond to the note frequency and intensity of the song, whereas the y axis will correlate closely to the changes in pitch throughout the song. These features will be demonstrated over time throughout each song with respect to time; the z-axis.

\begin{figure}
     \centering
     \begin{subfigure}[b]{.9\columnwidth}
         \centering
         \includegraphics[width=\textwidth,keepaspectratio]{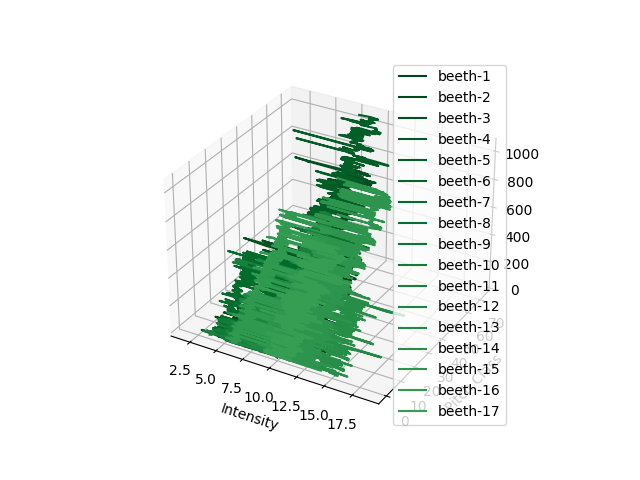}
         \caption{Beethoven's songs}
         \label{fig:graph1}
     \end{subfigure}
     \hfill
     \begin{subfigure}[b]{.9\columnwidth}
         \centering
         \includegraphics[width=\textwidth,keepaspectratio]{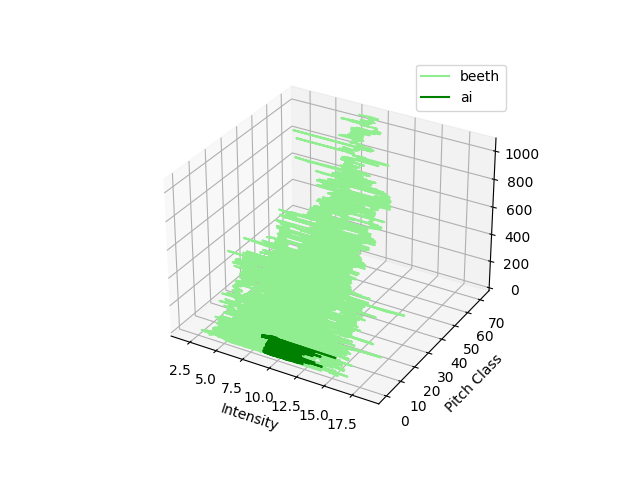}
         \caption{Our improved model in comparison to Beethoven's songs without scaling considered.}
         \label{fig:graph2}
     \end{subfigure}
     \begin{subfigure}[b]{.9\columnwidth}
     \hfill
         \centering
         \includegraphics[width=\textwidth,keepaspectratio]{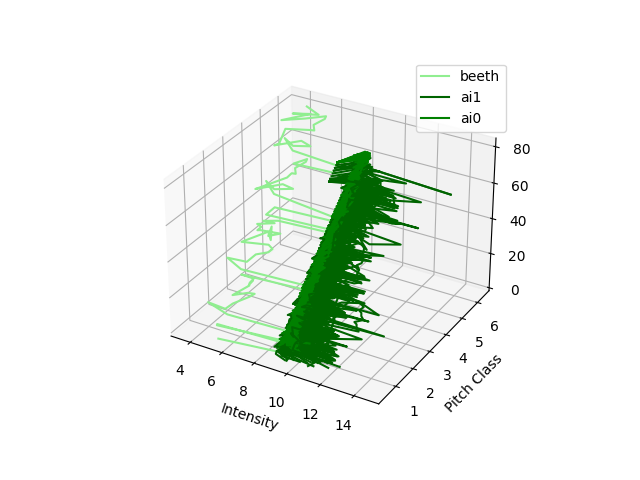}
         \caption{Beethoven's songs less than 120 segments in length and both models, AI-0 (original) and AI-1 (Neural Base).}
         \label{fig:graph3}
     \end{subfigure}
     \caption{Figure \ref{fig:graph1} shows the differences between Beethoven and the dataset, Figure \ref{fig:graph2} shows the differences between our model and Beethoven and Figure \ref{fig:graph3} shows a more in depth comparison with samples of similar song length.}
     \label{fig:biggraph}
\end{figure}

In order to obtain a closed form for each dimension, we must compress our feature space. Given that the features of each axis are correlated to the same metric, we reduced the $x$ and $y$ coordinate by taking the average of each row. To reduce the spread of these values, we also normalized this average. This allowed us to maintain the value of each entry and high resolution as depicted in Figure \ref{fig:biggraph}. Future studies can explore additional techniques to optimize this metric.

Beethoven's trend can be seen in Figure \ref{fig:graph1}. Each of his songs were placed into the graph under the same constraints, however, each song had a slightly different color of green. To ensure a fair comparison was made, we reduced the number of songs to 120 segments or less and added 50 samples from our improved model into Figure \ref{fig:graph3} along with 50 samples from the baseline system.

A promising trend revealed similar changes in pitch class, however, there are differences between note intensity. This is likely a result of the generation sequence favoring shorter note durations and can be explored in future studies. To our knowledge, however, obtaining such information has never been achievable. Additionally, we can demonstrate how our Neural Base approach out performed the previous iteration in Figure \ref{fig:graph3}.

Only one of Beethoven's samples is less than 150 segments. When the graph is extended to 200 segments, 2 of Beethoven's songs appear. Even at 350 segments, 5 of Beethoven's works appear. At 500 segments, 10 of Beethoven's works are included in the comparison. In turn, this comparison is limited to Figure \ref{fig:graph2}, where the segment threshold is less than 120 segments in order to maintain high resolution. We did not prune our samples from this comparison, as this could potentially skew the results.

\subsection{Survey}

Subjective performance was assessed using a general survey of our generated songs in comparison to samples found within the dataset. This survey is critical to music performance, as music is subjective. While an objective mapping of these generated songs can be obtained in comparison to Beethoven, the songs must also sound composed by a human and pleasant to the ear. The results from this survey support the system's ability to generate music while maintaining the compositional integrity. The following section includes details of this survey, including candidate song selection, survey participants/composition and the results.

\subsubsection{Candidate Sample Selection}

The 2 iterations of our system were used to generate one candidate sample. The first iteration demonstrates how the system performed without any additional attempts to remedy segment sparsity. The second iteration is the Neural Base implementation. One song was selected from each group to reduce the number of questions in the survey.

Each system iteration generated 5 pre-samples for candidacy. We selected 3 volunteers familiar with the classical music genre to blindly listen to each batch of samples and determine the most pleasing song from each iteration. This was done to ensure each model had the best chance in the survey while eliminating as much bias as possible.

Each volunteer voted for what they interpreted was the best song from each iteration, factoring commonly used metrics like song pleasantness and structure. While pleasantness is more noticeable to the untrained ear, structure is a more subtle metric that defines how a song progresses over time. Most notably in classical music, there is a beginning, middle, end and finale section -each of which are quite distinct as noted by our experts. The highest scoring songs from each iteration were selected to represent the neural network in our survey.

\subsubsection{Survey Composition}

To assess our system's ability to produce human-like compositions, a survey was designed to compare each iteration with real samples. One popular piano sonata from Mozart, Vivaldi and Beethoven were randomly selected to ensure the highest chance of survey participant recognition -making it exceedingly more difficult for our samples to perform as well as these accomplished composers. Given that these songs are considered among the most popular for each composer, attributes such as song pleasantness and structure were not factored into the selection.

The survey consisted of 5 music samples, asking participants the following questions:

\begin{enumerate}
    \item Who composed Sample $i$? [Option 1: Human, Option 2: Computer]
    \item Identify the composer of Sample $i$ (Leave blank if you do not know)
\end{enumerate}

These questions were chosen specifically to help gauge the participant's familiarity with the classical genre as well as give some insights to how our songs compared to human composers. Additionally, participants were asked to tell us the name of the composer if they were familiar with the composition.

Another factor considered in the survey is the time required for each participant to complete the survey. Some of the compositions from these esteemed composers could easily exceed 10 minutes in length. Thus, we reduced the samples to the initial 30 seconds of audio and eliminated the white noise leading up to the beginning of each song. On average, participants took half of this allotted time and no participant stated that they did not have enough audio to make their decision.

\subsubsection{Results}

After sampling 27 participants with a wide range of classical music experience, the average score came out to 60.74 with a median score of 60, with a standard deviation of 27.45 -which is approximately 7 points higher than the value of a single question (20 points). The margin of error in these values computed to 5.28.

Out of the survey participants, 7 scored on the extreme ends (2 scored 0, 5 scored perfectly). The majority of participants scored within the 40-80 point range, and one participant scoring 20. Additionally, one participant attempted to identify the composer for the first sample but did not do so correctly. The other participants did not attempt to identify the composer in any of the samples. 37\% of participants accurately labeled Vivaldi's work as human, 74\% accurately labeled Mozart and approximately 78\% of participants labeled Beethoven's work accurately. By comparison, our baseline iteration was accurately labeled as computer generated 74\% of the time, however, our neural-base implementation was accurately labeled computer generated 41\% of the time. Conclusively, 26\% of participants could not distinguish our baseline iteration from human composers and 59\% of participants believed our neural-base iteration was from a human composer.

\section{Conclusion and Future Works}

Knowledge bases were quite useful considering the limited amount of computation available at the beginning of AI research. As advances in hardware continued, researchers began exploring new and exciting deep learning models that have efficiently solved many more challenging tasks. Thus, the applications of knowledge bases became limited in modern practice.

In this paper, we introduce Neural Base, a knowledge base with a new architecture to facilitate a new form of inference for intelligence duplication. We demonstrate the Neural Base technique through the generation of melodies in the style of Beethoven. The effectiveness is assessed using a new methodology to measure performance based on the authentication of the generated music.  We also perform a qualitative analysis to demonstrate the Neural Base's ability to generate pleasant melodies -even though it is not immediately optimized for this.

By automating the rule set generation, we can significantly leverage a knowledge base to optimize intelligence duplication tasks with the addition of a neural network component. Our neural network component learns the placement style between individual building blocks and establishes the rules necessary to order this information based upon the dataset composer. With this strategy, we can add more information to this Neural Base without impeding our ability to retain the style learned from the original composer -mitigating segment sparsity in ID research.

In future works, we will integrate a generalized composer classifier and composer-specific classifier neural network architectures to provide more insights into the subjective aspect of music generation. This study concludes that testing the capacity of creative reasoning within ID networks should have multiple facets (in-person, heuristic, objective and subjectively automated). Our Neural Base approach gives researchers the ability to explore AI applications with unique creative reasoning; a feat that has yet to be explored.

\bibliography{aaai23.bib}

\end{document}